\documentclass[11pt]{article}
\usepackage[margin=1in]{geometry}
\usepackage{setspace}
\usepackage{times}
\usepackage{graphicx}%
\usepackage{multirow}%
\usepackage{amsmath,amssymb,amsfonts}%
\usepackage{booktabs}%
\usepackage{caption}
\usepackage{hyperref}
\usepackage{fancyhdr}
\usepackage{chngcntr}

\begin{document}

\begin{center}

{\Large \textbf{An Automated Radiomics Framework for Postoperative Survival Prediction in Colorectal Liver Metastases using Preoperative MRI}}\\[1em]

Muhammad Alberb$^{1,2,*}$, 
Jianan Chen$^{3}$, 
Hossam El-rewaidy$^{4}$, 
Paul Karanicolas$^{5}$, 
Arun Seth$^{6}$, \\
Yutaka Amemiya$^{6}$, 
Anne Martel$^{1,2}$, 
Helen Cheung$^{7,8}$\\[0.3em]

\footnotesize
$^{1}$ Department of Medical Biophysics, University of Toronto, Toronto, ON, Canada; 
$^{2}$ Physical Sciences Platform, Sunnybrook Research Institute, Toronto, ON, Canada; 
$^{3}$ UCL Cancer Institute, University College London, London, UK; 
$^{4}$ Systems and Biomedical Engineering Department, Cairo University, Giza, Egypt; 
$^{5}$ Department of Surgery, University of Toronto, Toronto, ON, Canada; 
$^{6}$ Department of Laboratory Medicine and Pathobiology, University of Toronto, Toronto, ON, Canada; 
$^{7}$ Sunnybrook Health Sciences Centre, Toronto, ON, Canada; 
$^{8}$ Department of Medical Imaging, University of Toronto, Toronto, ON, Canada\\[0.3em]

\small
$^*$Corresponding author: \texttt{muhammad.alberb@mail.utoronto.ca}

\end{center}

\begin{abstract}
\noindent \textbf{Purpose:} While colorectal liver metastasis (CRLM) is potentially curable via hepatectomy, patient outcomes remain highly heterogeneous. Postoperative survival prediction is necessary to avoid non-beneficial surgeries and guide personalized therapy. In this study, we present an automated AI-based framework for postoperative CRLM survival prediction using pre- and post-contrast MRI.

\noindent \textbf{Methods:} We performed a retrospective study of 227 CRLM patients who had gadoxetate-enhanced MRI prior to curative-intent hepatectomy between 2013 and 2020. We developed a survival prediction framework comprising an anatomy-aware segmentation pipeline followed by a radiomics pipeline. The segmentation pipeline learns liver, CRLMs, and spleen segmentation from partially-annotated data, leveraging promptable foundation models to generate pseudo-labels. To support this pipeline, we propose SAMONAI, a prompt propagation algorithm that extends Segment Anything Model to 3D point-based segmentation. Predicted pre- and post-contrast segmentations are then fed into our radiomics pipeline, which extracts per-tumor features and predicts survival using SurvAMINN, an autoencoder-based multiple instance neural network for time-to-event survival prediction. SurvAMINN jointly learns dimensionality reduction and survival prediction from right-censored data, emphasizing high-risk metastases. We compared our framework against established methods and biomarkers using univariate and multivariate Cox regression.

\noindent \textbf{Results:} Our segmentation pipeline achieves median Dice scores of 0.96 (liver) and 0.93 (spleen), driving a CRLM segmentation Dice score of 0.78 and a detection F1-score of 0.79. Accurate segmentation enables our radiomics pipeline to achieve a survival prediction C-index of 0.69.

\noindent \textbf{Conclusion:} Our results demonstrate the potential of integrating segmentation algorithms with radiomics-based survival analysis to deliver accurate, automated, and interpretable CRLM outcome prediction.
\end{abstract}

\section{Introduction}\label{sec:introduction}
Colorectal cancer is the second leading cause of cancer-related death worldwide~\cite{Sung2021GLOBOCAN}. The liver is the predominant site of colorectal cancer metastases~\cite{Wang2023CRLM} and surgical resection remains the only potentially curative treatment~\cite{Dong2023Surg}. Nevertheless, patient outcomes remain highly heterogeneous~\cite{Spolverato2013SurgicalOutcome}. Postoperative survival prediction is essential to avoid non-beneficial surgeries and guide personalized therapy, ultimately improving survival~\cite{Margonis2022PrecisionSurgery}. 

Conventional clinical and molecular biomarkers lack sufficient predictive power~\cite{Tian2024PrognosticFactors,Margonis2022PrecisionSurgery}, motivating imaging-based risk scores~\cite{Cheung2019TTE}. In multifocal CRLM, prognosis is often driven by aggressive tumors~\cite{Sebagh2016HeteroCRLM}. However, existing imaging scores typically focus on the largest tumor, potentially overlooking multifocality~\cite{Chen2021AMINN}.

Moreover, these scores require manual tumor segmentation, which is labor-intensive and prone to inter-rater variability~\cite{Horvat2024RadiomicsLimitations, Cheung2019TTE}. Automating segmentation via supervised training of deep learning models is limited by the scarcity of fully-annotated datasets~\cite{Rayed2024DLLimitations}. One solution is to complete missing annotations using promptable foundation models, such as Segment Anything Model (SAM)~\cite{Kirillov2023SAM} and its medical adaptation, MedSAM~\cite{Ma2024MedSAM}. 
SAM, however, only works on 2D images and therefore requires slice-wise prompting. Although SAM2~\cite{Ravi2024SAM2} and SAM3~\cite{Carion2025Sam3} overcome this limitation by treating slices as video frames, recent studies show that this approach does not necessarily outperform SAM~\cite{Sengupta2025SAMvsSAM2,Zhu2024MedicalSAM2,Buyukpatpat2025SAMLungComp}. 
MedSAM additionally requires box prompts, which are unsuitable for concave structures, and may suffer, as a result of medical finetuning, from reduced generalizability to unseen protocols~\cite{Zhang2024SamTTA}. 
Thus, a depth-aware extension of SAM without medical finetuning, box prompts, or slice-wise prompting is desirable.

Segmentation errors, particularly those corresponding to detecting false lesions in non-hepatic regions, can degrade downstream survival prediction ~\cite{Chen2021AMINN}. This highlights the importance of developing anatomy-aware segmentation models. Furthermore, automated methods in prognostic studies often utilize post-contrast images only~\cite{Chen2019AEGMM, Chen2021AMINN}. In clinical practice, radiologists typically evaluate both pre- and post-contrast scans to detect lesions and assess their malignancy~\cite{Chernyak2018LIRADS}. This discrepancy underscores the importance of incorporating pre-contrast features into prognostic models.

Following segmentation, a typical radiomics workflow includes feature extraction, feature selection, and modeling. However, decoupling feature selection from prediction can result in losing predictive signals~\cite{Liyanage2020JointSelectClassif}. To mitigate this, one multiple instance learning (MIL) approach jointly performs feature compression and 3-year survival prediction by aggregating tumor-level representations into patient-level outcomes ~\cite{Chen2021AMINN}. However, this approach formulates survival prediction as a binary classification task and excludes censored cases. Additionally, it averages tumor representations, which can obscure contributions from high-risk tumors.

The objective of this study is to investigate methods for improving postoperative survival prediction in CRLM. We hypothesize that leveraging pre- and post-contrast MRI to model high-risk tumors through an end-to-end automated framework outperforms conventional methods and existing biomarkers. To this end, we propose a two-stage framework comprising segmentation and radiomics. Our contributions are:
\begin{itemize}
  \item A segmentation pipeline that leverages foundation models to learn from partial labels. 
  \item SAMONAI, an algorithm that extends SAM to 3D point-based segmentation.
  \item SurvAMINN, an autoencoder-based MIL neural network for joint dimensionality reduction and survival prediction, accommodating censoring and emphasizing high-risk tumors.
\end{itemize}

\section{Materials and Methods}\label{sec:methods}
\subsection{Dataset}\label{subsec:dataset}
This retrospective study used an institutional dataset of 227 CRLM patients (139 males, 61 years, 31--84; 88 females, 61 years, 29--88) who underwent gadoxetate-enhanced MRI prior to curative-intent hepatectomy between January 2013 and December 2020. The institutional research ethics board granted approval and waived the requirement for informed consent. 

Studies were obtained as per institutional clinical protocols. Patients received 3D axial T1-weighted imaging (TE$\approx$1.5 ms, TR$\approx$3.0 ms, flip angle$\approx$10 degrees) in the pre-contrast and hepatobiliary phases (20 minute delay). A standard 10 mL intravenous dose of gadoxetate was administered at 1.0 mmol/mL. Studies were performed on 1.5-T or 3.0-T magnets with an eight-channel body phased array coil covering the liver.

An abdominal radiologist with 13 years of experience manually segmented all CRLMs with longest axial diameters $\geq 10\,\text{mm}$ in both pre- and post contrast images using ITK-SNAP~\cite{Yushkevich2006ITKSnap}. Segmented tumors per patient ranged from 1 to 12 with multifocal CRLM present in 130 patients. To reduce annotation cost, the liver and spleen were manually segmented for 20 post-contrast images only, which were reserved exclusively for evaluation.

Mortality status, survival time, and demographic data, including age and sex, were collected. 82 patients had known postoperative survival time (median 28 months), and the rest were right-censored. Risk scores, including Fong score~\cite{Fong1999Fong} and target tumor enhancement~\cite{Cheung2019TTE}, were calculated. For 60 patients, DNA sequencing was conducted following~\cite{Seth2021MutationCRLM}, and gene mutational statuses were recorded, including APC, TP53, KRAS, and NRAS. 

Exclusion criteria and data splitting are illustrated in Fig.~\ref{fig:dataset}, with more details available in supplemental material.

\begin{figure*}[htb]
\centering
\includegraphics[width=\textwidth]{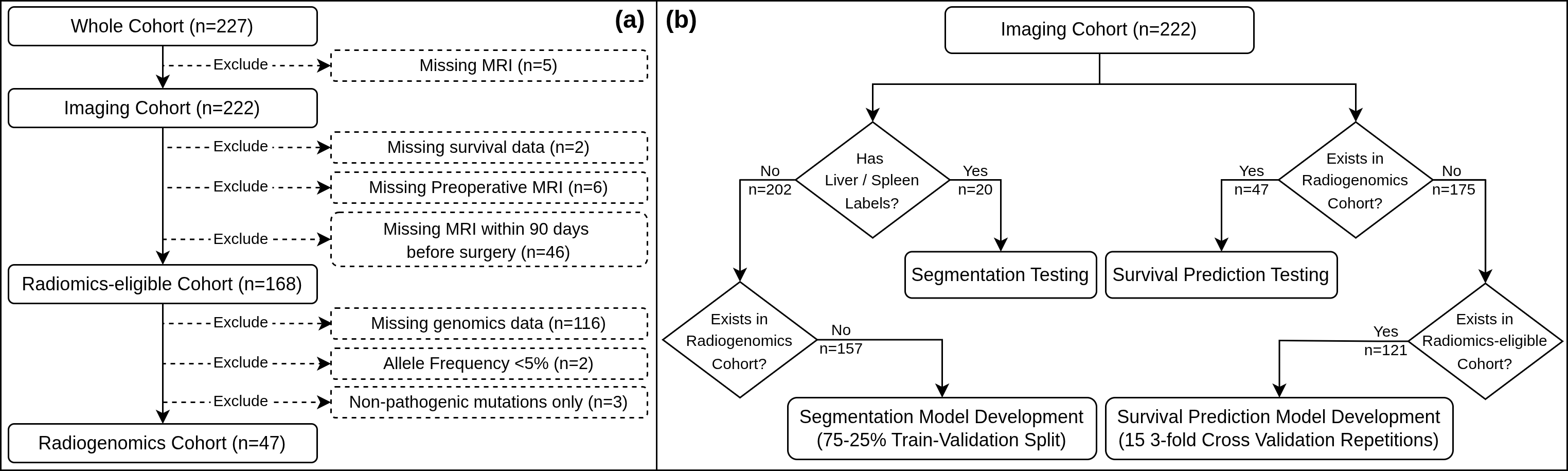}
\caption{Dataset processing: (a) cohort definitions and exclusion criteria, (b) patient-level data splitting for training, validation, and testing. Patients with manual liver and spleen annotations were held out from segmentation model development, while patients with available genomics data were held out from both segmentation and survival prediction models development to be exclusively used for testing.}\label{fig:dataset}
\end{figure*}

\subsection{Framework Overview}
We propose a fully automated framework for CRLM survival prediction from MRI, consisting of a segmentation pipeline followed by a radiomics pipeline.

The main objective of the segmentation pipeline is to delineate CRLMs. To enforce anatomical plausibility and suppress extra-hepatic predictions, liver segmentation is required. Additionally, spleen segmentation is incorporated to reduce ambiguity, as automated models often confuse it with the liver~\cite{Hossain2023LiverSpleenAmbiguity}. However, due to the limited availability of manual liver and spleen annotations in our dataset, cases with these labels were reserved for evaluation, with only CRLMs labels available during training. To enable learning from partial labels, the segmentation pipeline leverages a promptable foundation model during training to provide pseudo-labels for missing targets. To support this approach and ensure generalizability, we introduce SAMONAI, an algorithm that extends SAM to 3D segmentation.

Following segmentation, the radiomics pipeline extracts per-tumor features then models time-to-event survival using SurvAMINN, our MIL network that emphasizes high-risk tumors. During inference, radiomics are extracted from predicted segmentations, enabling end-to-end automation.

\subsection{Segmentation Pipeline}
\subsubsection{Segmentation with Partial Labels}
\begin{figure}[htb]
    \centering
    \includegraphics[width=0.9\textwidth]{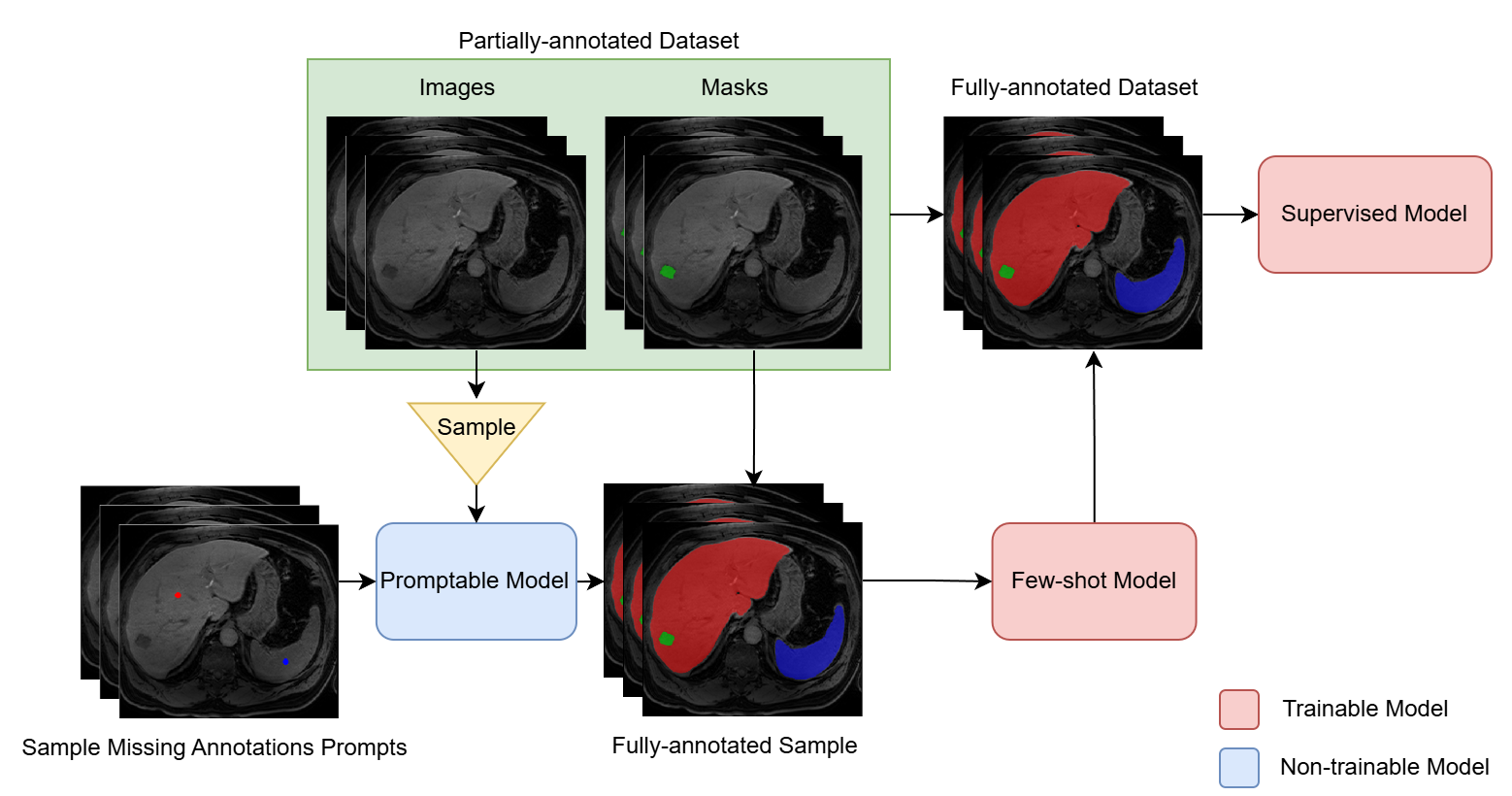}
    \caption{Segmentation pipeline: a promptable foundation model is used to complete partially-annotated masks for a small subset. A few-shot UNETR model is trained on the subset then used to complete the annotations for all cases selected for segmentation model development. Finally, A supervised UNETR model is retrained on the entire training dataset with completed labels.}
    \label{fig:segmentation_pipeline}
\end{figure}

Our segmentation pipeline, shown in Fig.~\ref{fig:segmentation_pipeline}, is designed to enable training with partially-labeled datasets. First, a promptable model is applied to segment the liver and spleen in a small subset. These segmentations are used as pseudo-labels to fine-tune a UNETR~\cite{Hatamizadeh2022UNETR} initialized with a pretrained vision transformer~\cite{xu20253DINO} and augmented with an adapter~\cite{Chen2022ViTAdapter}. To minimize manual effort during training, this stage is limited to 15 training and 5 validation samples. Subsequently, this finetuned UNETR is applied to the entire training dataset to generate liver and spleen segmentations. Finally, another training round is conducted to incorporate all CRLM labels. This pipeline is applied independently to pre- and post-contrast MRI. Preprocessing and training are detailed in supplemental material and follow~\cite{Alberb2025LiverDie}.

\subsubsection{Segmentation with Prompts}
\begin{figure}[htb]
    \centering
    \includegraphics[width=0.9\textwidth]{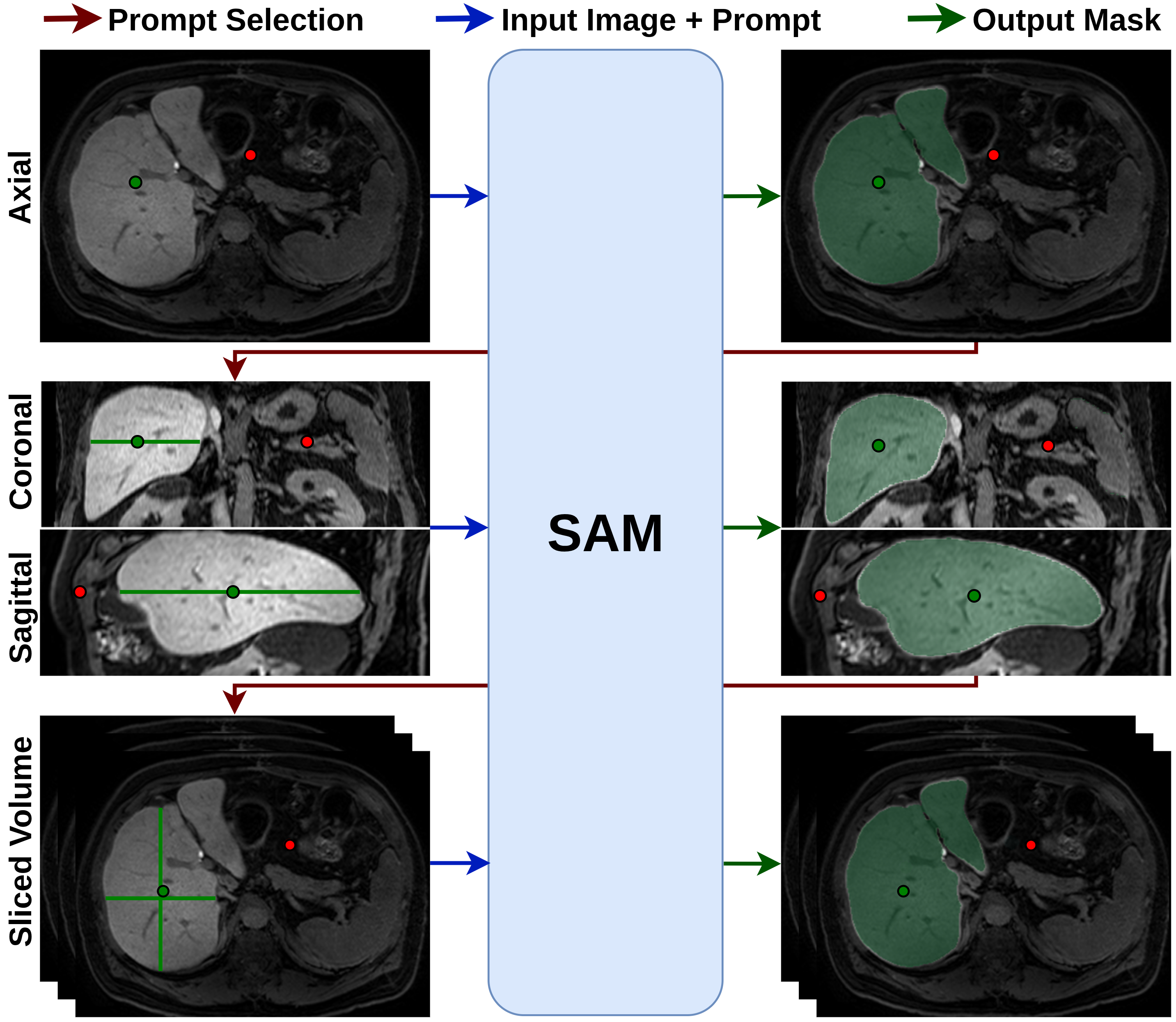}
    \caption{Overview of SAMONAI applied to the liver: 3D objects are segmented from single points by propagating prompts from one view to another. Positive and negative points are displayed in green and red, respectively.}
    \label{fig:samonai}
\end{figure}

During training, our segmentation pipeline requires a promptable model for unlabeled structures. To ensure robustness across protocols, we introduce SAMONAI, a prompt propagation algorithm, shown in Fig.~\ref{fig:samonai}, that extends SAM to volumetric segmentation.

Initially, the user provides prompts in a single slice from any view and SAM segments it. Consequently, positive lines are projected onto intersecting slices from other views. From each of these views, the slice with the longest positive line is segmented by selecting one positive point from the projected line and one negative point from background points along this line. 

Segmenting three slices from orthogonal views produces positive lines spanning the object. SAMONAI selects positive prompts from these lines and negative prompts beyond a bounding box enclosing the lines, after which SAM segments all slices in parallel.

To incorporate 3D awareness, this process is applied to all three views and predictions are averaged. Computational cost is preserved by sampling slices at $\frac{1}{3}$ density and interpolating skipped ones. Predictions are binarized with an adaptive threshold $T = \mu + 2\sigma$, where $\mu$ and $\sigma$ are the logits mean and standard deviation.

To select optimal prompts, SAMONAI follows three criteria derived from the observation that SAM performs better when prompts match the object’s average intensity, lie near its center, and reside in homogeneous regions. For a given point $p$ from candidate points $P$ in image $I$, an intensity criterion $C_{i}$ is defined as the absolute intensity difference between $p$ and the median of $P$:

\begin{equation}
C_{i}(p) = \left| I(p) - \text{median}_{q \in P}(I(q)) \right|
\label{eq:intensity}
\end{equation}

\noindent A location criterion $C_{l}$ is defined as the Euclidean distance between $p$ and the centroid of $P$:

\begin{equation}
C_{l}(p) = \left\| p - \frac{1}{|P|} \sum_{q \in P} q \right\|
\label{eq:location}
\end{equation}

\noindent A homogeneity criterion $C_{h}$ is defined as the standard deviation of intensities within an $11 \times 11$ neighborhood $N_{11 \times 11}$ centered around $p$:

\begin{equation}
C_{h}(p) = \sqrt{ \frac{1}{|N|} \sum_{r \in N_{11 \times 11}(p)} \left( I(r) - \mu_p \right)^2 },
\label{eq:homogeneity}
\end{equation}

\noindent where $\mu_p$ is the mean intensity of $N_{11 \times 11}$. The total cost $C_{t}$ is a weighted sum of equations~\eqref{eq:intensity}, \eqref{eq:location}, and \eqref{eq:homogeneity}, normalized over $P$:

\begin{equation}
C_{t}(p) = \alpha \cdot \hat{C}_{i}(p) + \beta \cdot \hat{C}_{l}(p) + \gamma \cdot \hat{C}_{h}(p),
\label{eq:total}
\end{equation}

\noindent where:
\begin{itemize}
    \item $\hat{C}_{i}(p)$, $\hat{C}_{l}(p)$, and $\hat{C}_{h}(p)$ are the min-max normalization of $C_{i}(p)$, $C_{l}(p)$, and $C_{h}(p)$.
    \item $\alpha$, $\beta$, and $\gamma$ are weights empirically set to 1, 1, and 2, respectively.
\end{itemize}

\noindent For negative points, we exclude the lowest 10\% of intensities to avoid selecting a background point. Finally, the optimal prompt is determined by minimizing equation~\eqref{eq:total}.

\subsection{Radiomics Pipeline}
\begin{figure}[htb]
    \centering
    \includegraphics[width=0.9\textwidth]{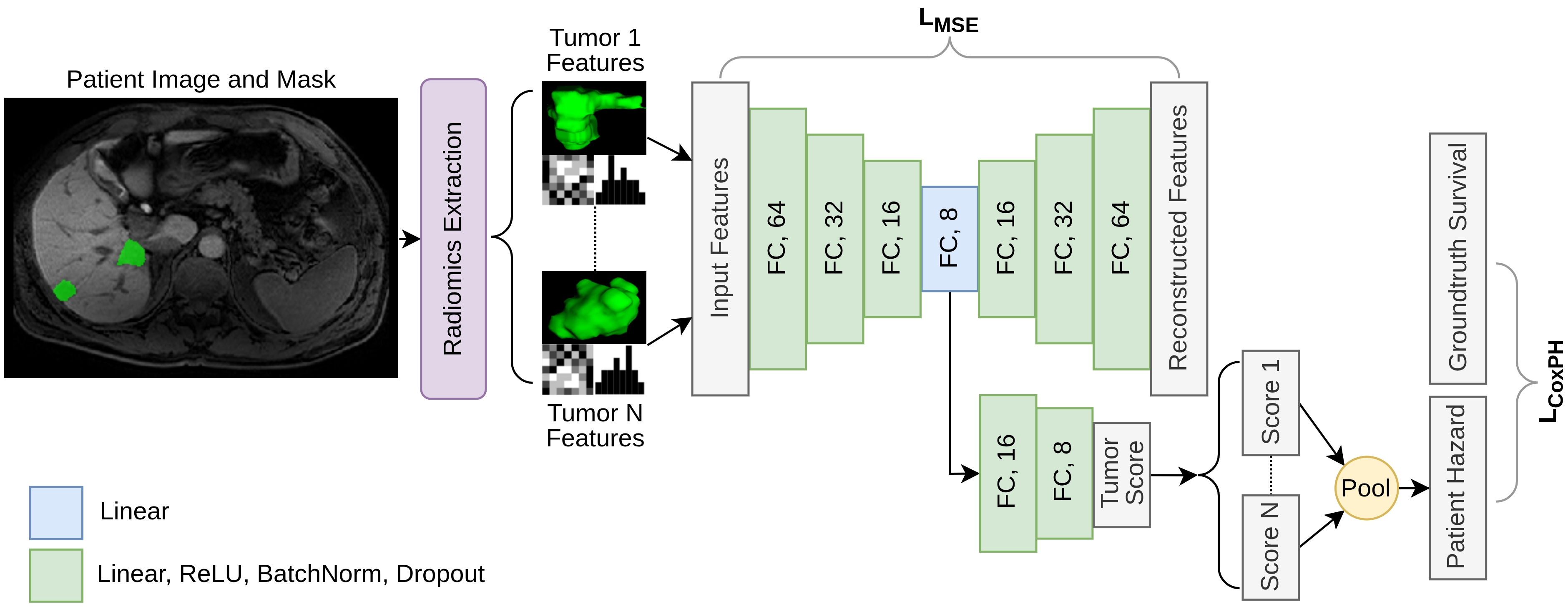}
    \caption{Radiomics pipeline: per-tumor radiomics are extracted. An autoencoder compresses features and a multiple instance regressor connected to the autoencoder's bottleneck predicts per-tumor risk scores and pools them into patient hazards. The autoencoder and regressor are jointly trained using a weighted sum of mean squared error $\mathcal{L}_{\text{\tiny MSE}}$ and cox proportional hazard loss $\mathcal{L}_{\text{\tiny CoxPH}}$.}
    \label{fig:radiomics_pipeline}
\end{figure}

After segmenting CRLMs, our radiomics pipeline, shown in Fig.~\ref{fig:radiomics_pipeline}, extracts per-tumor radiomics then performs simultaneous dimensionality reduction and survival prediction.

\subsubsection{Radiomics Extraction}
Radiomics are extracted using PyRadiomics~(v3.1.0)~\cite{VanGriethuysen2017PyRadiomics}. Predicted segmentations with longest diameter $\leq$ $1^{st}$ percentile of the training set are excluded, reducing instability. Volumes are resampled to isotropic $2\,\text{mm}$ spacing using B-spline interpolation. Outliers beyond $[\mu - 3\sigma, \mu + 3\sigma]$ are clipped, where $\mu$ and $\sigma$ are the intensity mean and standard deviation, respectively. Intensities are z-score normalized, scaled by 100, and discretized into bins of width 5. Statistical, shape, and textural features are extracted, as detailed in supplemental material, and normalized using two-step normalization~\cite{Chen2021AMINN}.

\subsubsection{Survival Prediction}
We propose SurvAMINN, a dense neural network comprising an autoencoder for dimensionality reduction and a MIL regressor for survival prediction from multifocal CRLM. SurvAMINN extends AMINN~\cite{Chen2021AMINN} by modeling time-to-event, handling censored data, and emphasizing high-risk tumors. The autoencoder is trained to embed per-tumor features and reconstruct them, by minimizing mean squared error:

\begin{equation}
\mathcal{L}_{\text{\tiny MSE}} = \frac{1}{T} \sum_{t=1}^{T} | x_t - \hat{x}_t |^2,
\label{eq:MSE}
\end{equation}

\noindent where $x_t$ and $\hat{x}_t$ are the input and reconstructed features, respectively, and $T$ is the number of tumors. 

The MIL regressor is trained to predict tumor scores from embeddings and pool them into patient hazards. To emphasize high-risk tumors, we define patient hazard $\eta_p$ as the logarithm of sum of exponentials (LSE) of tumor scores $\{\eta_1, \eta_2, \ldots, \eta_T\}$:

\begin{equation}
\eta_p = \log\left( \sum_{t=1}^{T} \exp(\eta_t) \right),
\label{eq:LSE}
\end{equation}

This behaves as a soft maximum, letting more aggressive tumors dominate predictions while accounting for cumulative contributions of multiple lesions and preserving differentiability for optimization. The regressor is optimized using Cox proportional hazard loss:

\begin{equation}
\mathcal{L}_{\text{\tiny CoxPH}} = - \sum_{p: \delta_p = 1} \left( \eta_p - \log \sum_{\hat{p} \in \mathcal{R}(T_p)} e^{\eta_{\hat{p}}} \right),
\label{eq:cox}
\end{equation}

\noindent where $\delta_p$ indicates mortality, $T_p$ denotes survival time, and $\mathcal{R}(T_p)$ are patients at risk at $T_p$. The total loss $\mathcal{L}_t$ is a weighted combination of equations~\eqref{eq:MSE} and \eqref{eq:cox}:

\begin{equation}    
\mathcal{L}_t = (1 - \alpha) \cdot \mathcal{L}_{\text{\tiny MSE}} + \alpha \cdot \mathcal{L}_{\text{\tiny CoxPH}},
\label{eq:total_loss}
\end{equation}

\noindent where $\alpha = \frac{\text{current epoch}}{\text{total epochs}} \in [0, 1]$ is a scheduling parameter that gradually shifts training from reconstruction to survival prediction, improving stability. 

\section{Results}\label{sec:results}
\subsection{Segmentation}

\begin{figure*}[htb]
    \centering
    \includegraphics[width=\textwidth]{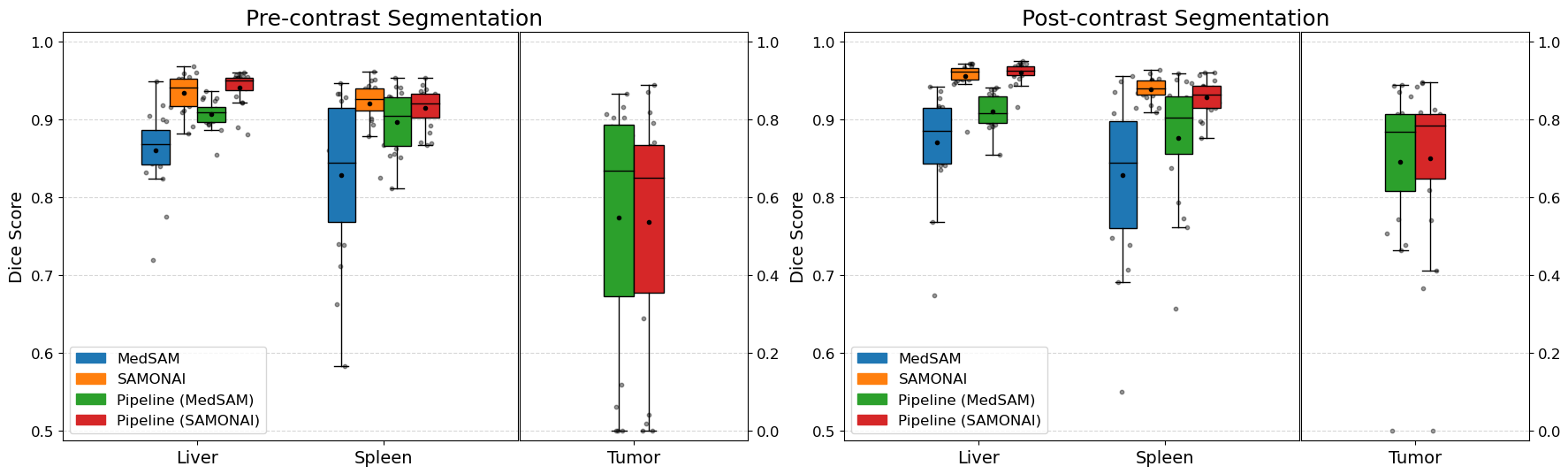}
    \caption{Segmentation box plots. MedSAM, in blue, and SAMONAI, in orange, denote prompt-based segmentation performance on missing labels. Pipeline (MedSAM), in green, and Pipeline (SAMONAI), in red, denote final automated segmentation performance after integrating each promptable model into our segmentation pipeline. 
    Black dots inside boxes represent mean Dice scores over patients, and gray dots outside boxes represent individual patients beyond the interquartile range.}
    \label{fig:segmentation_results}
\end{figure*}

Segmentation algorithms were evaluated using 20 fully-annotated cases, as shown in Fig.~\ref{fig:dataset}. Due to the high annotation cost, liver and spleen were only annotated in post-contrast images. To evaluate pre-contrast predictions, annotations were registered to pre-contrast images using affine followed by deformable registration via Elastix~\cite{Klein2010Elastix}.

Fig.~\ref{fig:segmentation_results} reports segmentation performance under two evaluation settings. First, we directly compare SAMONAI to MedSAM on missing labels. SAMONAI consistently outperforms MedSAM, with median Dice improvements of 7\% (liver) and 8\% (spleen) on pre-contrast MRI, and 8\% (liver) and 10\% (spleen) on post-contrast MRI.

Second, we assess the impact of each of them on our segmentation pipeline. SAMONAI improves the pipeline's median Dice scores by 8\% and 6\% (liver) and by 9\% and 5\% (spleen) on pre- and post-contrast MRI, respectively, yielding median Dice scores of 0.95 (liver) and 0.92 (spleen) on pre-contrast images, and 0.96 (liver) and 0.93 (spleen) on post-contrast images.  Tumor segmentation achieves median Dice scores of 0.65 and 0.78 for pre- and post-contrast, respectively, with comparable performance for SAMONAI and MedSAM, as both contribute mainly to missing labels.

Moreover, we evaluate the tumor detection performance of our segmentation pipeline, as shown in table~\ref{tab:detection_results}. As expected due to the enhanced tumor visibility, the post-contrast results surpass the pre-contrast ones, with a 0.79 F1-score. Nevertheless, pre-contrast performance remains satisfactory, with a 0.71 F1-score,  despite the inherent challenges in distinguishing tumors without contrast enhancement. A visual example of segmentation quality is provided in Supplementary Fig.~\ref{fig:qualitative}. These results highlight the potential of pre-contrast predictions to contribute meaningfully to survival analysis.

\begin{table}[htb]
\centering
\caption{Tumor detection results of the SAMONAI-based segmentation pipeline}
\begin{tabular}{@{}lllllll@{}}
\toprule
Contrast & Precision & Recall & F1-score & TP & FP & FN \\
\midrule
Pre-contrast  & 0.674 & 0.756 & 0.713 & 31 & 15 & 10 \\
Post-contrast & 0.720 & 0.878 & 0.791 & 36 & 14 & 5 \\
\bottomrule
\end{tabular}
\label{tab:detection_results}
\begin{center}
\vspace{-0.25cm}
\footnotesize \textbf{Note:} TP = true positives, FP = false positives, FN = false negatives.
\end{center}
\end{table}

\subsection{Survival Prediction}
In evaluating our radiomics pipeline, unless explicitly stated otherwise, we utilize predicted segmentations from both pre- and post-contrast MRI. We conducted four sets of experiments. 

\begin{figure}[htb]
    \centering
    \includegraphics[width=0.9\textwidth]{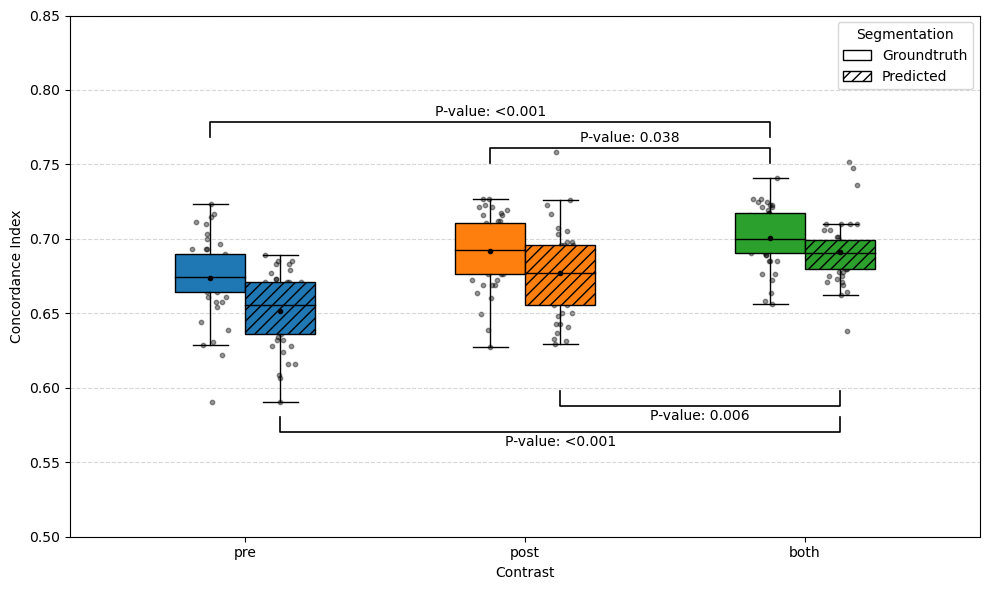}
    \caption{Survival prediction box plot showing C-index when using pre-contrast (pre), post-contrast (post), and combined contrast phases (both). Black dots inside boxes represent mean C-index over runs, and gray dots outside boxes represent individual runs beyond the interquartile range. Solid boxes correspond to ground-truth manual segmentations and striped boxes correspond to predicted segmentations. Statistical significance, assessed by Wilcoxon rank-sum test's p-value, confirms that combining pre- and post-contrast MRI yields better performance than using either phase individually.}
    \label{fig:radiomics_results}
\end{figure}

First, we compare survival prediction using pre- and post-contrast phases individually versus jointly. Figure~\ref{fig:radiomics_results} presents the results using predicted versus manual segmentations. Post-contrast segmentations yield better performance than pre-contrast alone. However, combining both phases consistently improves performance, as confirmed by Wilcoxon rank-sum test~\cite{Mann1947WilcoxonRankSum}. Although replacing ground-truth with predicted segmentations slightly degrades C-Index (0.023 for pre-contrast, 0.015 for post-contrast, and 0.010 for both), the pipeline retains strong predictive power. Kaplan-meier curves and log-rank tests, shown in Supplementary Fig.~\ref{fig:kaplan_meier}, reflect our pipeline's capability to stratify patients into low- and high-risk groups.

Second, we compare our pooling strategy with alternative approaches. Table~\ref{tab:pooling_results} presents results obtained by defining patient hazard as the score of the largest tumor, the mean tumor score, or the maximum tumor score. Survival prediction based on the highest-risk tumors, as captured by max pooling, is more accurate than relying on the largest tumor alone or averaging across all tumors. LSE pooling further improves performance by accounting for contributions from multiple high-risk tumors, enhancing stability and accuracy.

\begin{table}[htb]
\centering
\caption{Comparison of pooling strategies}
\begin{tabular}{@{}llll@{}}
\toprule
Pooling & C-Index & HR (95\% CI) & p \\
\midrule
mean        & 0.556 & 1.114 (1.040--1.216) & 0.759   \\
largest     & 0.594 & 1.263 (1.147--1.391) & 0.618   \\
max         & 0.623 & 1.568 (1.441--1.784) & 0.162   \\
LSE         & 0.691 & 1.929 (1.890--1.980) & $<$0.001 \\
\bottomrule
\end{tabular}
\label{tab:pooling_results}
\begin{center}
\vspace{-0.25cm}
\footnotesize \textbf{Note:} HR = hazard ratio, CI = confidence interval, p = Wald statistic P-value.
\end{center}
\end{table}

Third, we compare SurvAMINN to independent dimensionality reduction and survival prediction stages, as shown in Table~\ref{tab:baselines}. For feature reduction, we evaluate four approaches: no reduction (all features), K-best selection, principal component analysis (PCA), and minimal-redundancy-maximal-relevance (mRMR)~\cite{Peng2005mRMR}. For each reduction method, we train two models: support vector machines (SVM) for 3-year survival binary classification, and random survival forest (RSF)~\cite{Ishwaran2008RandomSurvivalForests}. Although SVM generally outperforms RSF, its performance is less stable, likely due to the reduced sample size after excluding censored cases. In contrast, SurvAMINN jointly learns dimensionality reduction and survival prediction using the entire dataset, enabling it to consistently outperform machine learning approaches.

\begin{table}[htb]
\centering
\caption{Comparison of SurvAMINN against common machine learning methods}
\begin{tabular}{@{}lllll@{}}
\toprule
Model & Reduction & C-Index & HR (95\% CI) & p \\
\midrule
SVM  & None     & 0.659 & 1.781 (1.693--1.876) & 0.002     \\
SVM  & KBest    & 0.647 & 1.745 (1.652--1.842) & 0.005     \\
SVM  & PCA      & 0.667 & 1.907 (1.793--2.036) & 0.004     \\
SVM  & mRMR     & 0.634 & 1.568 (1.494--1.652) & 0.011     \\
RSF  & None     & 0.632 & 1.447 (1.368--1.531) & 0.072     \\
RSF  & KBest    & 0.637 & 1.543 (1.453--1.653) & 0.056     \\
RSF  & PCA      & 0.634 & 1.475 (1.412--1.540) & 0.012     \\
RSF  & mRMR     & 0.631 & 1.446 (1.373--1.523) & 0.052     \\
\multicolumn{2}{l}{SurvAMINN} & 0.691 & 1.929 (1.890--1.980) & $<$0.001 \\
\bottomrule
\end{tabular}
\label{tab:baselines}
\begin{center}
\vspace{-0.25cm}
\footnotesize \textbf{Note:} HR = hazard ratio, CI = confidence interval, p = Wald statistic P-value. None refers to using all extracted features without dimensionality reduction.
\end{center}
\end{table}

Fourth, we compare SurvAMINN to existing clinical and molecular biomarkers, as shown in table~\ref{tab:biomarkers}. Although gene mutations, including APC, TP53, KRAS, and NRAS, are acquired after surgery from resected samples, they fail to accurately predict survival. Despite the improvement in the C-index when combining SurvAMINN with these biomarkers in a multivariate Cox model, SurvAMINN remains the only predictor with a strong association with survival, as revealed by the hazard ratio (HR), confidence interval (CI), and p-value. Finally, we conduct a randomized test of SurvAMINN, as shown in Supplementary Fig.~\ref{fig:random_test}, to further confirm the statistical significance of the achieved C-index.

\begin{table}[htb]
\centering
\caption{Comparison of SurvAMINN against existing clinical and genomic biomarkers using univariate and multivariate Cox regression}
\begin{tabular*}{\textwidth}{@{\extracolsep\fill}lcccccc}
\toprule
& \multicolumn{3}{@{}c@{}}{Univariate} & \multicolumn{3}{@{}c@{}}{Multivariate} \\\cmidrule{2-4}\cmidrule{5-7}
Biomarker & C-Index & HR (95\% CI) & p & C-Index & HR (95\% CI) & p \\
\midrule
\textbf{\textit{Clinical factors}} & & & & \multirow{13}{*}{0.751} & & \\[-3pt]
\quad Sex  & 0.476 & 0.93 (0.57--1.51) & 0.768 & & 1.16 (1.12--1.20) & 0.691 \\
\quad Age  & 0.525 & 1.34 (0.81--2.22) & 0.258 & & 1.54 (1.50--1.58) & 0.164 \\
\textbf{\textit{Risk scores}} & & & & & & \\[-3pt]
\quad Fong~\cite{Fong1999Fong}  & 0.545 & 1.17 (0.74--1.85) & 0.500 & & 0.91 (0.88--0.94) & 0.741 \\
\quad TuEn~\cite{Cheung2019TTE} & 0.538 & 1.21 (0.81--1.80) & 0.347 & & 1.12 (1.09--1.17) & 0.737 \\
\textbf{\textit{Genomic mutations}} & & & & & & \\[-3pt]
\quad APC  & 0.583 & 0.82 (0.50--1.34) & 0.418 & & 0.59 (0.58--0.61) & 0.127 \\
\quad TP53 & 0.502 & 1.28 (0.80--2.07) & 0.306 & & 1.71 (1.65--1.85) & 0.163 \\
\quad KRAS & 0.538 & 0.93 (0.57--1.54) & 0.785 & & 0.88 (0.86--0.90) & 0.668 \\
\quad NRAS & 0.500 & 1.17 (0.85--1.60) & 0.339 & & 0.85 (0.82--0.88) & 0.634 \\
\textbf{\textit{Ours}} & & & & & & \\[-3pt]
\quad SurvAMINN & 0.691 & 1.93 (1.89--1.98) & $<$0.001 & & 3.41 (3.17--4.02) & $<$0.001 \\
\bottomrule
\end{tabular*}
\label{tab:biomarkers}
\begin{center}
\vspace{-0.4cm}
\footnotesize \textbf{Note:} HR = hazard ratio, CI = confidence interval, p = Wald statistic P-value, TuEn = target tumour enhancement.
\end{center}
\end{table}

\section{Discussion}\label{sec:discussion}
This study presents a CRLM postoperative survival prediction framework from preoperative MRI based on an anatomy-aware segmentation pipeline. On one hand, our zero-shot promptable segmentation method, SAMONAI, achieved substantial liver and spleen segmentation accuracy. Our experiments show that SAMONAI can outperform medical image segmentation models in terms of accuracy and generalizability. Incorporating SAMONAI into our segmentation pipeline enables learning of liver, CRLMs, and spleen segmentation from partially-annotated data. Our anatomy-aware pipeline eliminates extra-hepatic tumors, improving detection performance for downstream survival prediction. 

In addition to the detection metrics described in table~\ref{tab:detection_results}, the predicted post-contrast segmentations were reviewed by the same radiologist who performed the manual segmentation described in section~\ref{subsec:dataset}. The radiologist's assessment revealed better detection performance than estimated. Our detection metrics count disconnected objects as independent tumors. Therefore, in case of ill-defined segmentation borders, this strategy overestimates false positives (FPs) and false negatives (FNs) by counting disconnected pixels as tumors. The radiologist's assessment revealed that 13 FPs and 6 FNs were not truly false, but rather single pixels or disconnected borders.

Moreover, another source of FPs is that CRLMs with longest diameter $\leq 1~cm$ are not annotated in the ground-truth. Our pipeline detected 7 small CRLMs that were counted as FPs. Benign lesions are also not included in the manual segmentations. Our pipeline segmented 6 benign lesions (3 cysts and 3 hemangiomas) leading to 6 additional FPs. While benign lesions minimally contribute to prognosis, they are not expected to severely impair our survival prediction pipeline, especially since SurvAMINN estimates patient hazards based on higher-risk tumors, as indicated in equation~\eqref{eq:LSE}.

Although removal of small objects in post processing reduced FPs from 29 to 14 and FNs from 8 to 5, many of the remaining errors were still attributable to the factors discussed above, primarily due to the conservative thresholds used for eliminating small objects. In fact, the radiologist's assessment revealed that only 3 FPs and 2 FNs were truly false, after neglecting those mentioned above. This suggests that the actual precision and recall exceed 0.93 and 0.95, respectively. 

Interestingly, the radiologist reported that 2 of the actual FPs corresponded to surgical resection margins in a patient who underwent pre-MRI hepatectomy. Notably, liver segmentation for this case was highly accurate (Dice score $\approx 0.97$). The third FP, observed in a different patient, was attributed to biliary duct dilation. These three, shown in Supplementary Fig.~\ref{fig:failure},  were the only truly incorrect FPs, likely due to the absence of similar abnormalities in training data.

These findings highlight the robustness of our segmentation pipeline. By eliminating the need for labor-intensive manual segmentation, reducing reliance on expert input, and minimizing inter-rater variability, our segmentation pipeline streamlines clinical workflows and paves the way for fully-automated postoperative survival prediction.

Our survival prediction results confirm our hypotheses. First, simultaneously utilizing pre- and post-contrast phases improves over individual ones, as shown in Fig.~\ref{fig:radiomics_results}. Second, high-risk tumors are more predictive of survival than average or large ones, as shown in table~\ref{tab:pooling_results}. LSE improvement over max pooling suggests that multiple high-risk tumors further impair survival. Third, table~\ref{tab:baselines} highlights that jointly learning dimensionality reduction and survival prediction improves over performing them independently, suggesting that SurvAMINN can preserve features that are crucial for survival prediction. Finally, SurvAMINN proves to be superior to existing clinical and molecular biomarkers shown in table~\ref{tab:biomarkers}.

Despite improved automation and predictive performance, our framework has several limitations. For instance, the modest gain from combining pre- and post-contrast MRI suggests insufficient complementary information and suboptimal late fusion, which was employed due to pre- and post-contrast misalignment. Incorporating more informative modalities and tumor-matching algorithms is expected to improve performance. Furthermore, exclusion of patients with missed tumors can introduce selection bias, highlighting the need to further improve detection sensitivity. Additionally, the segmentation pipeline can be sensitive to the samples selected for pseudo-labeling. Integrating active learning  and quality control strategies can enhance robustness. Also, our radiomics pipeline under-utilizes liver and spleen segmentations, which may have prognostic relevance if used for feature extraction. Finally, we aim to extend our framework to predict response to chemotherapy in non-surgical cohorts, which can enhance clinical utility by supporting treatment decision-making for patients who are not candidates for surgery, ultimately contributing to more personalized and effective therapy.

In conclusion, this study presents a fully automated CRLM postoperative survival prediction framework from pre- and post-contrast MRI. The framework integrates a liver, CRLMs, and spleen segmentation pipeline, powered by promptable foundation models, with a radiomics pipeline for joint dimensionality reduction and survival prediction based on high-risk tumors. Our experiments demonstrate that our framework can outperform existing machine learning methods and biomarkers, showing a potential to improve CRLM outcome prediction.

\bibliographystyle{vancouver}
\bibliography{bibliography}

\clearpage
\appendix
\renewcommand{\thefigure}{S\arabic{figure}}
\renewcommand{\thefigure}{S\arabic{figure}}
\setcounter{figure}{0}
\renewcommand{\thetable}{S\arabic{table}}
\setcounter{table}{0}
\raggedbottom

\begin{center}
{\LARGE \textbf{Supplemental Materials}}
\end{center}

\section{Framework Overview}
\begin{figure*}[htb]
    \centering
    \includegraphics[width=0.9\textwidth]{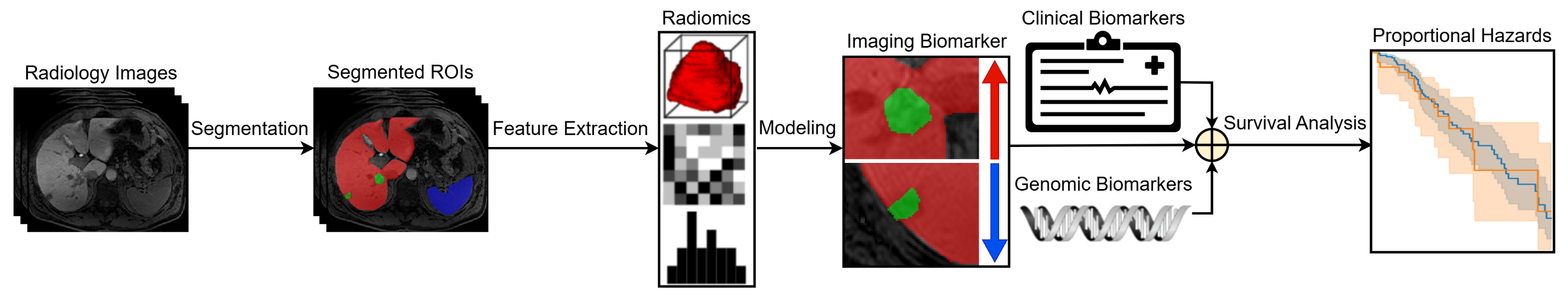}
    \caption{Framework overview: liver, tumors, and spleen are segmented and used to extract radiomic features. An imaging risk score is then predicted from radiomics and combined with other biomarkers to predict patient prognosis.}
    \label{fig:overview}
\end{figure*}

Our framework consists of a segmentation pipeline followed by a radiomics pipeline. An overview of the framework is shown in figure~\ref{fig:overview}.

\section{Dataset Description}
The dataset used for this study was internally acquired at our institution between January 2013 and December 2020. The retrospective cohort consists of 227 CRLM patients who have received pre- and post-contrast T1-weighted MRI with the hepatobiliary contrast agent gadoxetic acid (Primovist, Eovist). Ethical approval for this retrospective study was obtained from the institutional research ethics board, which waived the requirement for individual informed consent. All imaging data were de-identified and converted to NIfTI format prior to analysis.

Studies were obtained as per institutional clinical protocols. The patients received 3D axial T1-weighted imaging (TE$\approx$1.5 ms, TR$\approx$3.0 ms, flip angle$\approx$10 degrees) in the pre-contrast and hepatobiliary phases (20 minute delay).  For the contrast agent, a standard 10 mL intravenous dose of gadoxetate at 1.0 mmol/mL was administered.  Studies were performed on 1.5-T (GE Twinspeed\textsuperscript{\tiny{TM}}) or 3.0-T (Philips Achieva\textsuperscript{\tiny{TM}}) magnets with an eight-channel body phased array coil covering the liver.

The dataset also includes segmentations of CRLM tumors with a longest axial diameter $\geq 10\,\text{mm}$ for pre- and post-contrast images, overall mortality, and time to survival/death. The manual segmentations were conducted by an abdominal radiologist with 13 years of experience in abdominal segmentation. To validate unsupervised liver and spleen segmentation pipelines, the liver and the spleen were manually segmented by the same radiologist for 20 post-contrast images.

The number of segmented tumors per patient varies from 1 to 12 tumors, resulting in a total of 531 and an average of 2.4 tumors per patient. 130 patients have multifocal CRLM with an average of 3.4 tumors per patient. 82 patients have known postoperative survival times, with a median survival of 28 months. The rest of the patients are right-censored.

Furthermore, the dataset contains a few known prognostic variables. Specifically, the dataset includes demographic data such as age and sex, clinical biomarkers such as Fong score \cite{Fong1999Fong} and target tumor enhancement \cite{Cheung2019TTE}, as well as genomic biomarkers such as APC, TP53, KRAS, and NRAS mutational status available for a subset of 47 patients. DNA extraction and mutation analysis were performed following the same procedure described by \cite{Seth2021MutationCRLM}. Genomic mutations were filtered based on their allele frequency and ClinVar pathogenicity classification. The number of occurrences of the specified mutations per patient was used as a prognostic biomarker.

For this study, different inclusion criteria were defined for the segmentation and outcome prediction pipelines. For segmentation, all patients with available pre- or post-contrast MRI with tumor segmentation were included. For outcome prediction, additional inclusion criteria were applied, such as available clinical data including survival information and preoperative MRI acquisition $\leq 3\,\text{months}$ prior to the surgery date. For patients with multiple scans that met these criteria, the MRIs closest to the surgical date were selected for analysis. The subset with available manual liver and spleen segmentations was held out from the segmentation pipeline, and the subset with genomic data was held out from both pipelines to be used as independent testing sets.

\section{Experimental Setup} \label{sec:experimental_setup}
\subsection{Radiomics Extraction}
Extracted feature classes included first-order statistical, shape-based, gray level co-occurrence matrix (GLCM), gray level run length matrix (GLRLM), gray level size zone matrix (GLSZM), and gray level dependence matrix (GLDM) features. Table~\ref{tab:radiomics_summary} summarizes the number of features for each feature class.

\begin{table}[tb]
\centering
\caption{Summary of extracted radiomic features}
\begin{tabular}{@{}lll@{}}
\toprule
Class & Category & Features \\
\midrule
First-order      & Statistical & 18 \\
Shape-based      & Shape       & 14 \\
GLCM             & Textural    & 22 \\
GLRLM            & Textural    & 16 \\
GLSZM            & Textural    & 16 \\
GLDM             & Textural    & 14 \\
\multicolumn{2}{c}{Total} & 100 \\
\bottomrule
\end{tabular}
\label{tab:radiomics_summary}
\begin{center}
\vspace{-0.25cm}
\footnotesize \textbf{Note:} GLCM = gray level co-occurrence matrix, GLRLM = gray level run length matrix, GLSZM = gray level size zone matrix, GLDM = gray level dependence matrix.
\end{center}
\end{table}

\subsection{Evaluation Metrics}

To evaluate segmentation quality, we compute the Dice score for each of the liver, tumor, and spleen. We also evaluate tumor detection by treating tumor instances as detection targets. A predicted tumor is considered a true positive if it achieves a Dice score $\geq0.1$ with a ground-truth tumor. To enforce one-to-one matching, each ground-truth tumor is assigned to at most one predicted tumor, and vice versa, selected as the one with the highest Dice score above 0.1. Any remaining unmatched predicted tumors are counted as false positives, and unmatched ground truth tumors are counted as false negatives. Based on this strategy, we compute standard classification metrics: precision, recall, and F1-score. For outcome prediction, we use the concordance index (C-index) to assess the predictive performance of our method.

\subsection{Training Details}

\noindent \textbf{Segmentation.} Due to the high computational cost, segmentation models were trained once and validated using a randomly sampled 25\% hold-out validation set. Best validation checkpoints were evaluated on the test set. Models were trained for 1000 epochs with AdamW optimizer with a learning rate of $1 \times 10^{-4}$, and a weight decay of $1 \times 10^{-2}$, and batch size of 2, and validated every 2 epochs. Learning rate was scheduled with a linear warm-up for 10\% of iterations followed by cosine annealing.

For preprocessing, volumes were resampled to isotropic voxel spacing of 1\,mm. Intensity outliers were removed via 99.9th percentile clipping, and intensities were normalized to $[-1, 1]$. Empty background regions were cropped. Data augmentation included 3D flipping, rotations, intensity scaling, and shifting. A sliding window of $112 \times 112 \times 112$ was used during inference. For post-processing, extra-hepatic tumors as well as tumors with volume~$< 100~mm^3$ were removed. This threshold was empirically assigned and is much smaller than the smallest tumor in the training data. Training was conducted on a single NVIDIA A100 GPU. 

\noindent \textbf{Outcome Prediction.} Experiments were repeated 15 times using different 3-fold cross-validation splits. Since pre- and post-contrast tumor segmentations are unpaired, a late fusion strategy was used by aggregating pre- and post-contrast predictions. Models were trained for 250 epochs using AdamW with $4 \times 10^{-4}$ learning rate, $1 \times 10^{-3}$ weight decay, and 20\% dropout. Class imbalance was mitigated by sampling equal number of censored and uncensored patients per epoch.

\subsection{Statistical Analysis}

To examine the strength of association between individual variables and survival outcomes, we fit Cox Proportional Hazards models~\cite{Cox1972Cox} on z-score standardized variables and compare their Hazard Ratios (HR). We assess the statistical significance of our findings through multiple approaches. First, we use the Wilcoxon rank-sum test to compare performance between different input settings (i.e., pre/post-contrast vs. both contrasts)~\cite{Mann1947WilcoxonRankSum}. Second, we employ a randomization test by training and evaluating our model $1 \times 10^{3}$ different times on data with randomly shuffled survival labels to establish a null distribution. Similar to our actual setup, each random C-index was calculated by averaging scores from 15 different 3-fold cross-validation splits. 

Third, we stratify patients into high- and low-risk groups by median dichotomizing SurvAMINN hazards averaged over the 15 runs. The groups are compared using Kaplan–Meier curves and the log-rank test. Finally, we evaluate the robustness of HR across repeated runs with 95\% confidence intervals (95\% CI) of $1 \times 10^{3}$ bootstrapped HR estimates, and the two-sided \emph{P}-values are calculated using Wald test on log-transformed HRs.

\section{Additional Results} \label{sec:extra_figures}
\subsection{Segmentation}
\begin{figure*}[htb]
    \centering
    \includegraphics[width=0.85\textwidth]{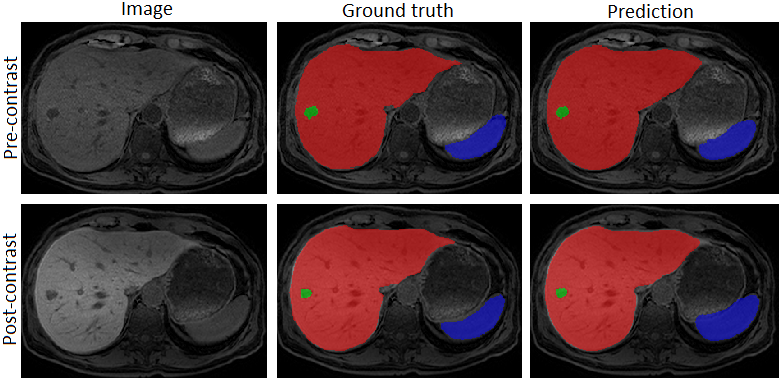}
    \caption{Qualitative assessment of the segmentation pipeline on a sample testing case.}
    \label{fig:qualitative}
\end{figure*}

\begin{figure*}[htb]
    \centering
    \includegraphics[width=0.85\textwidth]{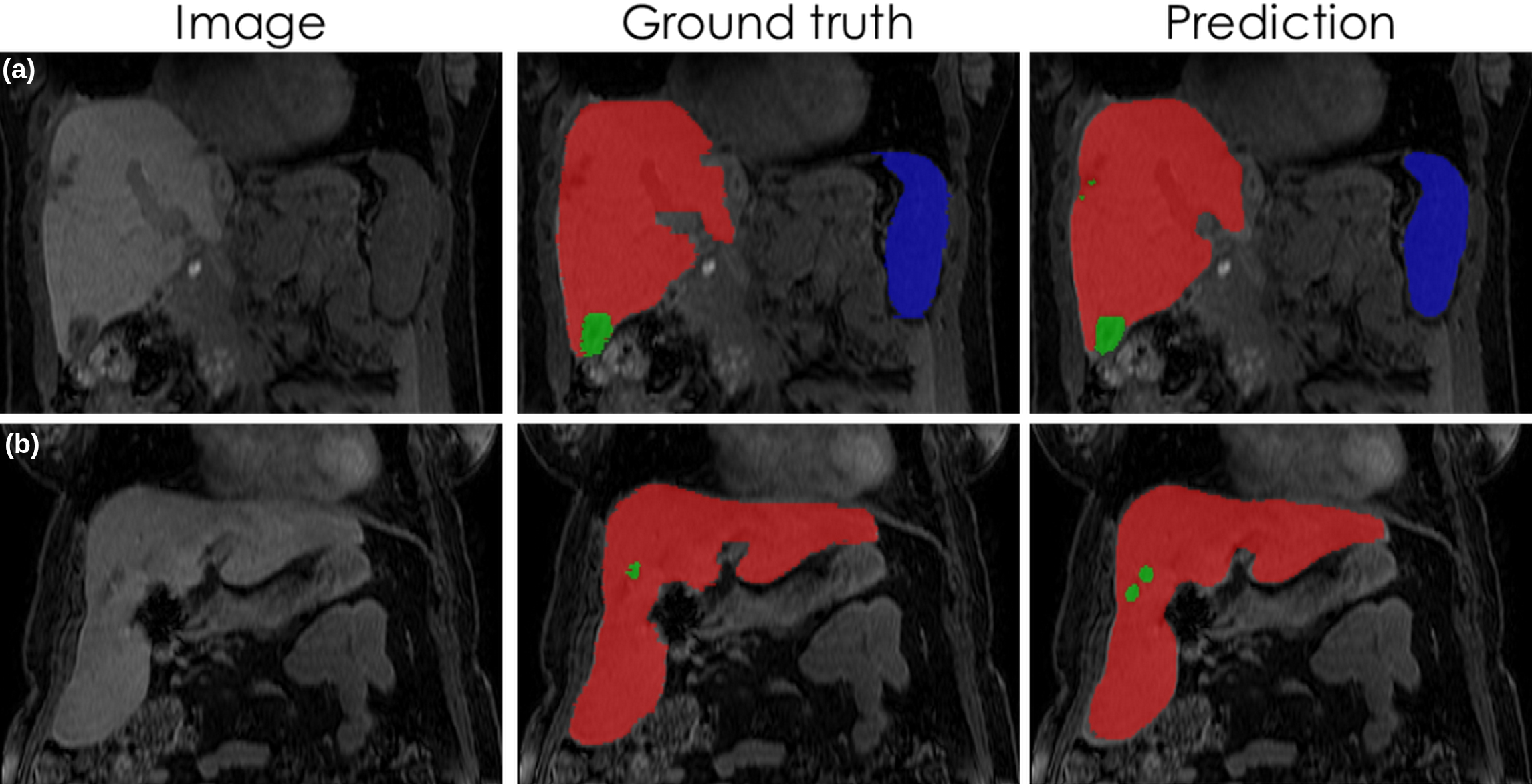}
    \caption{Failure analysis of the segmentation pipeline illustrating two testing cases with false-positive detections: (a) two FPs located near a surgical resection margin in a patient who underwent hepatectomy prior to MRI acquisition, and (b) one FP corresponding to biliary duct dilatation.}
    \label{fig:failure}
\end{figure*}

We provide a visual comparison of the predicted segmentations against the ground truth for a randomly selected test case, shown in Fig.~\ref{fig:qualitative}. The dice scores for the pre-contrast segmentation are 0.956 for liver, 0.742 for tumors, and 0.915 for spleen, while the post-contrast scores are 0.963, 0.786, and 0.922, respectively. The precise delineation observed in both contrast settings highlights the robustness of our pipeline and its potential to rival fully supervised approaches. We also present the two test cases in which the model produced false-positive tumor predictions (Fig.~\ref{fig:failure}). Both cases contain atypical abnormalities, suggesting that these errors may be attributable to the limited representation of such abnormalities in the training data.

\subsection{Outcome Prediction}
\begin{figure*}[tb]
    \centering
    \includegraphics[width=0.9\textwidth]{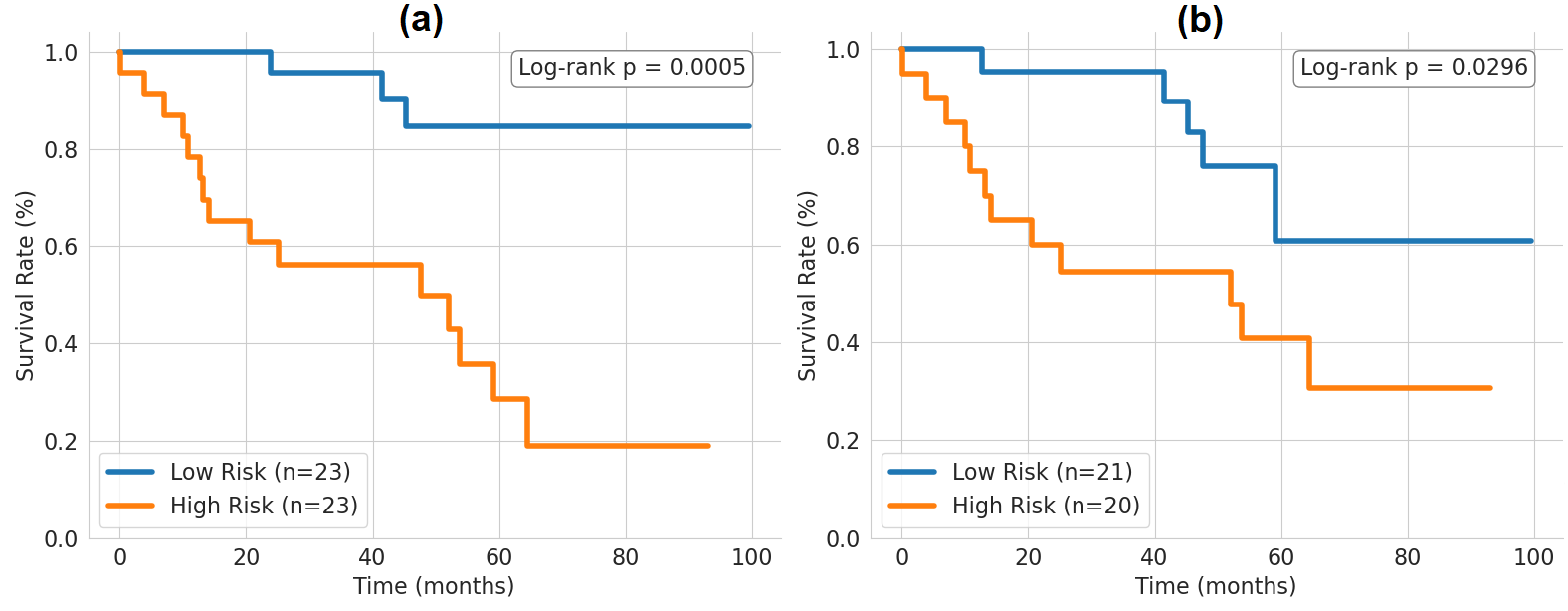}
    \caption{Kaplan–Meier curves of our outcome prediction pipeline using (a) ground truth segmentations and (b) predicted segmentations. Five patients, for whom no tumors were predicted, were excluded from the analysis in (b), as hazard scores could not be computed.}
    \label{fig:kaplan_meier}
\end{figure*}

\begin{figure}[htb]
    \centering
    \includegraphics[width=0.74\textwidth]{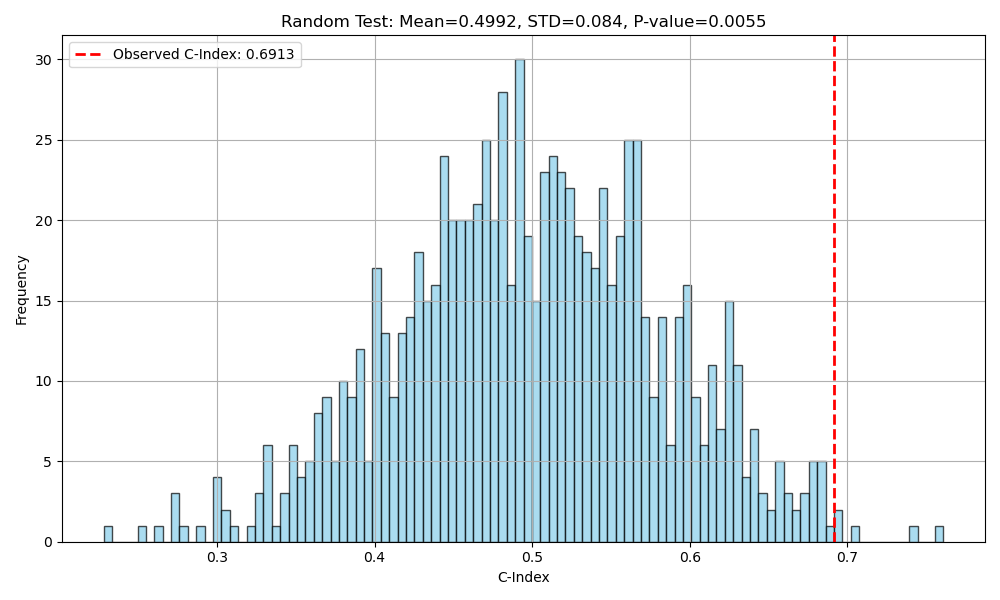}
    \caption{Randomized testing of SurvAMINN. Blue bars represent the C-index distribution when training and testing on randomly shuffled labels. Red line represents the observed score without randomization.}
    \label{fig:random_test}
\end{figure}

The kaplan-meier curves and log-rank tests, as shown in Fig.~\ref{fig:kaplan_meier}, demonstrate SurvAMINN's capability to stratify patients into high- and low-risk groups. This significance is further shown in the randomized test of SurvAMINN, as shown in Fig.~\ref{fig:random_test}

\end{document}